\documentclass[11pt]{article}
\usepackage{acl2023}
\usepackage{float}
\usepackage{times}
\usepackage{latexsym}
\usepackage{url}
\usepackage{hyperref}
\usepackage{booktabs}
\usepackage{graphicx}
\usepackage{color}
\usepackage{amsmath}
\usepackage[]{natbib}

\title{Joint Effects of Argumentation Theory, Audio Modality and Data Enrichment on LLM-Based Fallacy Classification}
\author{Hongxu Zhou ~\;~~\;~~ Hylke Westerdijk ~\;~~\;~~ Khondoker Ittehadul Islam\\[5pt]University of Groningen\\[5pt]
  {\tt \{h.zhou.16, h.p.westerdijk, k.i.islam\}}{\tt@student.rug.nl}\\
}
\date{\today}
\begin{document}
\maketitle

\begin{abstract}
This study investigates how context and emotional tone metadata influence large language model (LLM) reasoning and performance in fallacy classification tasks, particularly within political debate settings. Using data from U.S. presidential debates, we classify six fallacy types through various prompting strategies applied to the Qwen-3 (8B) model. We introduce two theoretically grounded Chain-of-Thought frameworks: Pragma-Dialectics and the Periodic Table of Arguments, and evaluate their effectiveness against a baseline prompt under three input settings: text-only, text with context, and text with both context and audio-based emotional tone metadata. Results suggest that while theoretical prompting can improve interpretability and, in some cases, accuracy, the addition of context and especially emotional tone metadata often leads to lowered performance. Emotional tone metadata biases the model toward labeling statements as \textit{Appeal to Emotion}, worsening logical reasoning. Overall, basic prompts often outperformed enhanced ones, suggesting that attention dilution from added inputs may worsen rather than improve fallacy classification in LLMs.{\footnote{Data and code available at \url{https://github.com/hongxuzhou/fallacy_detection}}}
\end{abstract}

\section{Introduction}

Fallacious arguments play a prominent role in political debates since antiquity.    In the pragma-dialectical theory of argumentation, fallacies can be defined as “speech acts that violate the rules of a rational argumentative discussion for assumed persuasive gains” \cite{ijcai2022p575}. An example of such a fallacy is the ad hominem, a rhetorical device employed to discredit an opponent’s arguments on the basis of their character or supposed shortcomings, instead of on the argument itself. Fallacies can be utilized by political figures to “skew rational judgement and influence public perception” \cite{Ifan_2022}, and by producers of fake news to spread misinformation \cite{Elena}, making their automatic detection in arguments socially relevant. 

For fallacy classification, previous authors have employed BERT-based models using the dialogue context, the fallacious argument snippet, the argument component, and argument relation as features \cite{ijcai2022p575}. Others have used Support Vector Machines and Neural Networks using everyday fallacious arguments as input features \cite{habernal-etal-2017-argotario}. \cite{mouchel2025logicalfallacyinformedframeworkargument} used GPT-4 for fallacy classification, using short argumentative texts as input. \cite{mancini-etal-2024-multimodal} used both audio and text in an encoder architecture. Multimodal fallacy detection with generative AI is thus far an unexplored domain.

\section{Related Work}

\subsection{Audio Modality}

Paralinguistics refers to the way audio features, such as pitch and tone, can convey more meaning than the literal text itself. This aspect has assisted in various multi-modal applications, including sarcasm detection \cite{tomar2023your}, emotion detection \cite{liang2024multi} etc., Recently, \cite{mestre2023augmenting} explored 13 audio features, such as contrast, bandwidth, and centroid, to determine whether a text is argumentative or not. The authors employed an independent encoder-decoder architecture for an audio module that integrated convolution layers and BiLSTM with a BERT-based textual model for argumentative classification. However, they reported a performance drop when incorporating audio data. Consequently, we opted not to include the audio feature as an independent audio module followed by concatenation.

A more complex audio feature involves extracting the arousal, dominance, and valence values of speech, also known as the emotional tone. \citet{wagner2023dawn} recently achieved state-of-the-art performance in classifying emotional tone from MS-Podcast dataset \cite{lotfian2017building}, which resembles our presidential debate dataset, as both environments are controlled and free from external noise. Notably, three of the six fallacy labels in our task—\textit{Appeal to Emotion}, \textit{Slogans}, and \textit{Ad Hominem}—are closely linked to emotional tone. To explore this connection, we utilized the wav2vec 2.0-large model to extract arousal, dominance, and valence values from audio recordings. Additionally, recent studies in multi-modal tasks, particularly emotion detection \cite{wu2025multi, zhao2022memobert}, have highlighted performance improvements when audio feature values are effectively integrated into prompts. Thus, we are taking this innovative approach in this fallacy classification task.

\subsection{Research Questions}
As a result, in this research project, we look into answering the following questions:
\begin{enumerate}
\item How do theory-informed prompting strategies affect LLM performance?
\item What is the impact of data enrichment (audio feature + context)?
\item When combining theoretical guidance and enrichment, what patterns emerge in model performance?
\end{enumerate}

\section{Task}

This project is centered around the MM-ArgFallacy2025 shared task, which is split in two challanges; •  Argumentative Fallacy Detection and •  Argumentative Fallacy Classification . In our study we have worked on the latter. For this sub-task, the goal was to take “a sentence, in the form of text or audio or both, extracted from a political debate, containing a fallacy.”, and classify it into the following six categories; ‘Appeal to Emotion’, ‘Appeal to Authority’, ‘Ad Hominem’, ‘False Cause’, ‘Slippery Slope’, and ‘Slogans’. The dataset of the shared task contains 1,278 labelled sentences from US presidential elections between 1960 and 2020, and includes audio fragments and the context of the sentence. 

\section{Methodology \& Experimental Setup}

\subsection{Data}
\begin{figure}[H]
    \centering
    \includegraphics[width=0.9\linewidth]{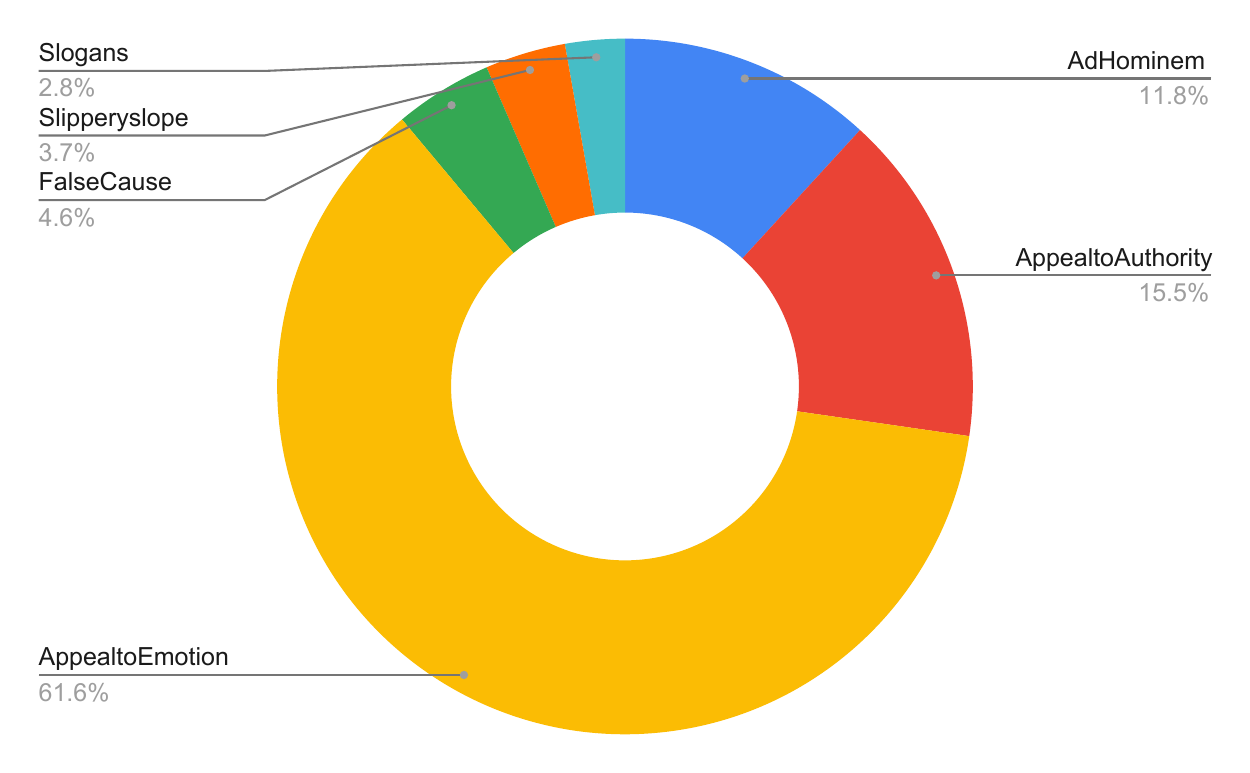}
    \caption{Class Imbalance Distribution of the original Mamkit Dataset.}
    \label{fig:class_imbalance}
\end{figure}

We utilize the MAMKit dataset \cite{manciniMultimodalFallacyClassification2024} which consists of 21,716 entries, each containing dates and several snippet IDs related to presidential and vice-presidential debates. Each snippet ID includes audio duration (start and end times), transcribed text, and a label corresponding to one of six fallacy types. Due to the imbalance in fallacy label distribution (illustrated in Figure \ref{fig:class_imbalance}), we selected 10 instances of each fallacy label for a validation set (60 instances total) and 20 instances for a test set (120 instances total) to adequately address our research question on class-balanced sets.

\subsection{Context: Date}
We also intend to examine how the context, specifically the date metadata from the mamkit dataset, influences fallacy classification. By integrating this date information, we expect the model to infer details about the speaker and the topic of each debate, which were not available in the market dataset.

\subsection{Audio Modality}
\begin{figure}[H]
    \centering
    \includegraphics[width=0.95\linewidth]{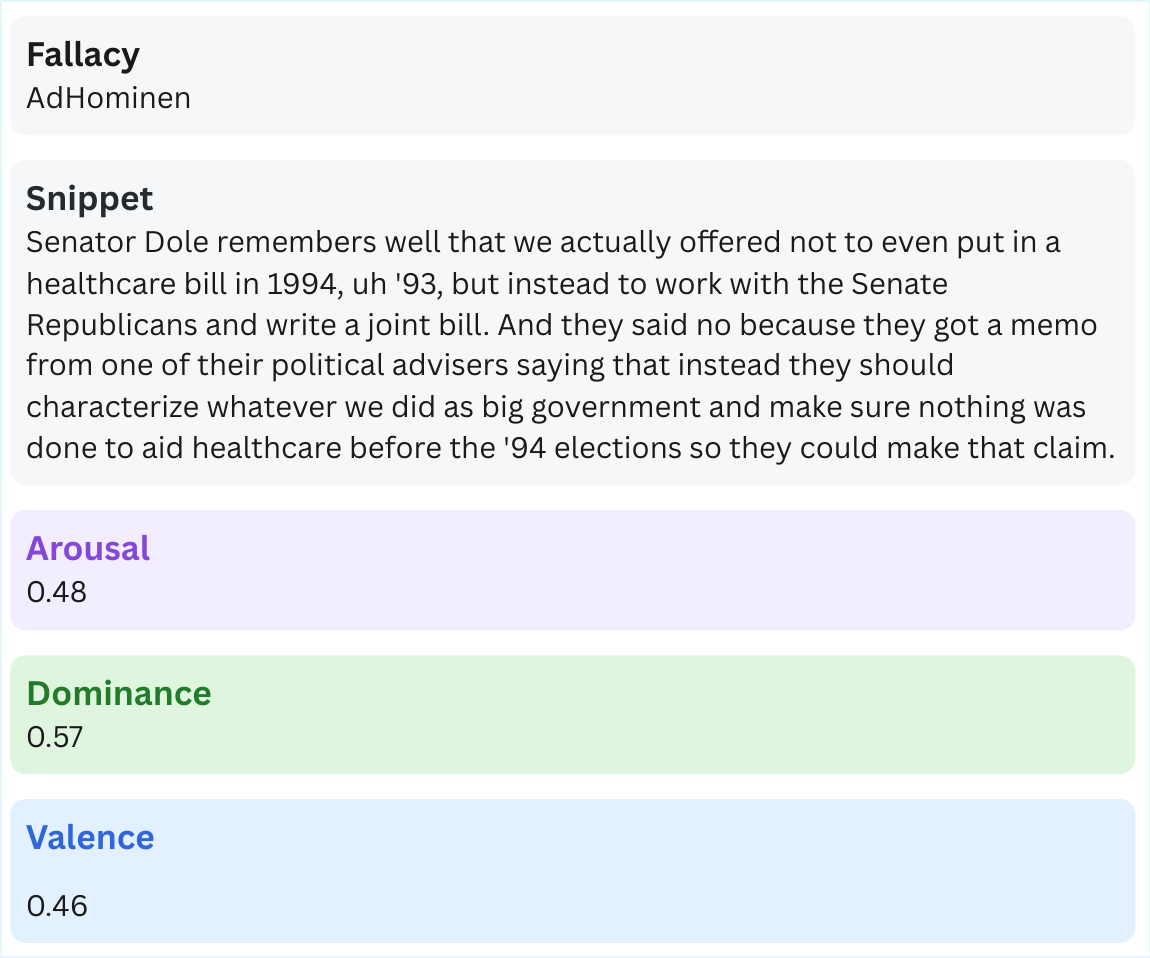}
    \caption{A Row Sample from the Validation Set comprising of the Fallacy Type, Classifiable Text, Feature Values of Each Emotional Tones.}
    \label{fig:row_sample}
\end{figure}

\begin{figure}[H]
    \centering
    \includegraphics[width=0.9\linewidth]{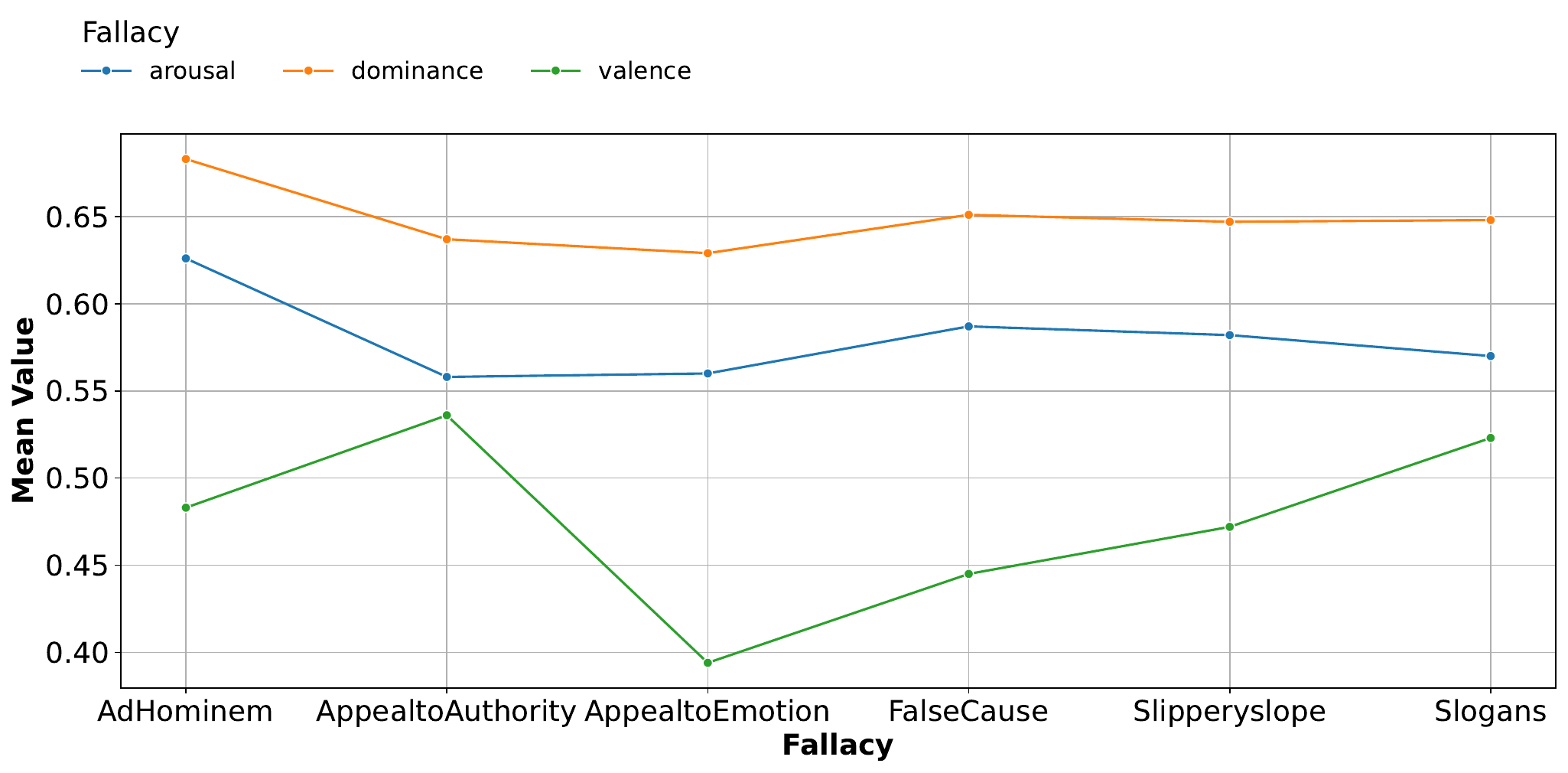}
    \caption{Mean Audio Feature Distribution of Arousal, Dominance and Valence across each Fallacy Type on the Validation Set. }
    \label{fig:emotional_tone_feature_distribution}
\end{figure}

We detect the arousal, dominance, and valence of each audio recording. The recordings are organized by date and snippet ID by the mamkit dataset. More formally, we load the pre-trained \verb|wav2vec 2.0-large| model from \citet{wagner2023dawn}. For each entry in our validation and test sets, we passed the audio through this model with \verb|resample| and \verb|verbose| set to True. Later, we preprocess the signal recommended by the author to receive final values ranging from -1 to 1 for each arousal, dominance, and valence. These scores were then appended to the respective rows in the dataset. We present a sample row including the emotional tone values and date context in Figure \ref{fig:row_sample}. Furthermore, we present the distribution of the mean emotional tone feature values across each fallacy type on the validation set in Figure \ref{fig:emotional_tone_feature_distribution}. Notably, since the dataset consists of US presidential debates, the mean dominance feature value is significantly higher compared to the other emotional tones. Interestingly, we observe that fallacy types have a more pronounced effect on valence compared to the other emotional tones.

\subsection{Experimental Setting}
As we also want to evaluate the reasoning capabilities of Large Language models in classifying fallacies, we chose the open-source \verb|Qwen-3| model, specifically the 8B parameterized version due to computational constraints. However, due to computational complexity, we only stick to the 8B parameterized version. In order to make the model generate non-monotonous tokens on reasoning, and analysis to reach the final answer, we set the temperature to $0.6$. To encourage the model to generate diverse responses during reasoning, we set the \verb|temperature| parameter to 0.6 and adjusted \verb|top-p| and \verb|top-k| settings to $0.95$ and $20$, respectively. Due to time complexity and resource limitations, we could not experiment with any other values of these parameters.

\subsection{Prompt Engineering}
In this section, we introduce and discuss the motivation and structure of our prompts: Basic prompt, Pragma-Dialectics (PD) \cite{vaneemerenFallaciesPragmadialecticalPerspective1987} Prompt, and Periodic Table of Arguments (PTA) \cite{wagemansHowIdentifyArgument2023} Prompt.

\subsubsection{Basic}
We consider this method as a baseline to compare our performance of the theoretical framework methods. As a result, we keep the structure of the prompt very basic. More formally, first, we provide the model with the instructions. Later, we provide the terminology and definition of each of the six fallacies. Finally, we also provide a general background of the task by explicitly depicting "\textit{All statements are taken from United States Presidential Debates between 1960 and 2020}"
\subsubsection{Pragma-Dialectics}
As the primary motivation of this theoretical framework is to guide the model to classify fallacy through eliminating possible fallacies which the statement violated to be as one, our prompt comprises instruction, task, and 10 PDA rules of their definition and violation criterion. To be specific on the task, for each statement, we ask the model to first identify the most significant violated rules. Later, we ask the model to classify the fallacy before providing a justification of why selecting this fallacy.

\subsubsection{Periodic Table of Arguments}
Our main goal of this theoretical framework is to guide the model to systematic reasoning before reaching the final answer. As a result, in the prompt, we first ask the model to break down every argument statement into (1) How are the parts arranged? (i.e., Form) (2) What kind of claims? (i.e., Substance) and (3) What connects the premise to the conclusion? (i.e., Lever). Then, we ask the model whether the argument is fallacious by explicitly providing rules such as "\textit{Alpha-FF needs causal/effect/sign levers}". Later, we provide systemic fallacy mapping where we explicitly map each fallacy to a specific constraint violation. Finally, to prioritize closely associated fallacies, we provide a clear hierarchy for example, “\textit{First check: Is there even an argument structure? (No → Slogan)}”

\subsubsection{Audio Modality}
To interpret the emotion tone values effectively, we categorized each value as low, moderate, or high, based on a heuristic approach: values below -0.33 are considered low, above 0.33 are high, and those around the threshold are moderate.  We also provide a definition of each emotion tone using the same heuristical approach. For arousal, we let the model know if the emotion is \textit{energetic}, \textit{lethargic}, or \textit{calm}. For dominance, if the emotional tone is \textit{assertive}, \textit{submissive}, or \textit{neutral in control} and for valence, if the emotional tone is \textit{positive}, \textit{negative}, or \textit{emotionally neutral}.

\subsection{Evaluation Metrics}
We tuned the model hyperparameters and our prompts on a validation set. We report model performance on our test set. While our primary evaluation metric is accuracy, we also analyze precision, recall, and F1-scores across different prompt types and conditions.

\section{Results}
This section presents the results of our fallacy classification experiments using large language models under three distinct prompting strategies: a basic prompt, a pragma-dialectical (PD) prompt leveraging structured reasoning based on argumentation theory, and a prompt inspired by the Periodic Table of Arguments (PTA). For each prompt type, we evaluate model performance across three input conditions: the base prompt alone, the prompt with additional context from the debate, and the prompt with both context and emotional tone metadata extracted from the audio. We analyze both quantitative metrics and qualitative reasoning outputs to assess how each setup influenced classifications. The results are presented per prompt type in the sections that follow, beginning with the basic approach.
\subsection{Base prompt}

\subsubsection{Performance Metrics}
\begin{table}[H]
\centering
\begin{tabular}{lccc}
\toprule
\textbf{Metric} & \textbf{B} & \textbf{C} & \textbf{C + A} \\
\midrule
Accuracy & \textbf{51\%} & 49\% & 46\% \\
Macro Precision & 58\% & 56\% & \textbf{66\%} \\
Macro Recall & \textbf{51\%} & 49\% & 46\% \\
Macro F1 & \textbf{49\%} & 47\% & 44\% \\
Weighted F1 & \textbf{49\%} & 47\% & 44\% \\
\bottomrule
\end{tabular}
\caption{Overall classification metrics for the base prompt across input conditions. Bold values indicate the highest score per row.}
\label{tab:pd-metrics}
\end{table}

The overall performance of the base prompt is relatively similar across the three input conditions, with a slight exception for the macro precision of context + audio.  

\subsubsection{Per-Class Performance}
\begin{table}[H]
\centering
\begin{tabular}{lccc}
\toprule
\textbf{Fallacy Type} & \textbf{F1-B} & \textbf{F1-C} & \textbf{F1-C + A} \\
\midrule
Appeal to Emotion & \textbf{46\%} & 45\% & 44\% \\
Appeal to Authority & \textbf{48\%} & 42\% & 45\% \\
Ad Hominem & \textbf{81\%} & 77\% & 78\% \\
False Cause & 41\% & \textbf{47\%} & 33\% \\
Slippery Slope & 9\% & 9\% & \textbf{10\%} \\
Slogans & \textbf{69\%} & 62\% & 53\% \\
\bottomrule
\end{tabular}
\caption{Per-class F1-score across input conditions for the base prompt. Bold values highlight the highest score per fallacy type.}
\label{tab:pd-f1-per-class}
\end{table}

\begin{table}[H]
\centering
\begin{tabular}{lccc}
\toprule
\textbf{Fallacy Type} & \textbf{F1-B} & \textbf{F1-C} & \textbf{F1-C + A} \\
\midrule
Appeal to Emotion & \textbf{33\%} & 31\% & 29\% \\
Appeal to Authority & 62\% & 54\% & \textbf{64\%} \\
Ad Hominem & \textbf{88\%} & 79\% & \textbf{88\%} \\
False Cause & 38\% & \textbf{41\%} & 38\% \\
Slippery Slope & 50\% & 50\% & \textbf{100\%} \\
Slogans & 80\% & \textbf{83\%} & 80\% \\
\bottomrule
\end{tabular}
\caption{Per-class Precision scores across input conditions for the base prompt. Bold values highlight the highest score per fallacy type.}
\label{tab:pd-f1-per-class}
\end{table}

\begin{table}[H]
\centering
\begin{tabular}{lccc}
\toprule
\textbf{Fallacy Type} & \textbf{F1-B} & \textbf{F1-C} & \textbf{F1-C + A} \\
\midrule
Appeal to Emotion & 80\% & 75\% & \textbf{95\%} \\
Appeal to Authority & \textbf{40\%} & 35\% & 35\% \\
Ad Hominem & \textbf{75\%} & \textbf{75}\% & 70\% \\
False Cause & 45\% & \textbf{55\%} & 30\% \\
Slippery Slope & 5\% & 5\% & 5\% \\
Slogans & \textbf{60\%} & 50\% & 40\% \\
\bottomrule
\end{tabular}
\caption{Per-class Recall across input conditions for the base prompt. Bold values highlight the highest score per fallacy type.}
\label{tab:pd-f1-per-class}
\end{table}

If we look at the performance metrics per fallacy type for the base prompt, we see more variety, both within input conditions and within fallacy type. The base prompt misses 95\% of all Slippery Slope fallacies, but the base prompt with context + audio achieves a precision score of 1.00 for this fallacy category. Conversely, snippets are often mistakenly classified as 'Appeal to Emotion', especially when audio is included. Furthermore, we see a relatively large discrepancy in recall scores between context and context + audio for the false cause fallacy. The inclusion of audio makes the model gravitate towards not classifying a snippet as containing a False Cause. 

\subsubsection{Confusion Matrices}

An interesting observation is that for the basic prompt, Slippery Slope snippets often get classified as ‘Appeal to Emotion’ and ‘False Cause’. When looking at the difference matrix for just context alone, we do not see a pronounced effect of added context for any particular fallacy type. Adding audio has a bad effect on the slogan detection of the model. 

\subsection{Pragma-Dialectics}
The pragma-dialectical prompt was designed to guide the model through a structured Chain-of-Thought (CoT) process grounded in the ten rules of critical discussion. This approach aimed to encourage rule-based reasoning in classifying the primary fallacy present in a given statement. In this section, we evaluate how well the model performed under the PD prompt across the 3 input conditions, and analyze both classification metrics and individual samples to understand the strengths and limitations of this method.

\subsubsection{Performance Metrics}
\begin{table}[H]
\centering
\begin{tabular}{lccc}
\toprule
\textbf{Metric} & \textbf{B} & \textbf{C} & \textbf{C + A} \\
\midrule
Accuracy & \textbf{53\%} & 49\% & 46\% \\
Macro Precision & 61\% & 58\% & \textbf{65\%} \\
Macro Recall & \textbf{53\%} & 49\% & 46\% \\
Macro F1 & \textbf{54\%} & 49\% & 46\% \\
Weighted F1 & \textbf{54\%} & 49\% & 46\% \\
\bottomrule
\end{tabular}
\caption{Overall classification metrics for the PD prompt across input conditions. Bold values indicate the highest score per row.}
\label{tab:pd-metrics}
\end{table}

Table~\ref{tab:pd-metrics} shows the model's overall classification performance under the pragma-dialectical prompt across three input settings: \textbf{B} (Base), \textbf{C} (Context), and \textbf{C + A} (Context + Audio). The base setup achieves the highest accuracy, macro recall, and F1-scores, suggesting that the model performs best when not provided with contextual input. Interestingly, while the context + audio condition leads to the highest macro precision, it performs worst on recall and F1, indicating that the model may become too focused in its predictions when influenced by emotional tone metadata. Overall, these results suggest that adding more contextual input does not consistently improve model performance, and in most cases may introduce noise or distract from the core problem of the statement.

\subsubsection{Per-Class Performance}
\begin{table}[H]
\centering
\begin{tabular}{lccc}
\toprule
\textbf{Fallacy Type} & \textbf{F1-B} & \textbf{F1-C} & \textbf{F1-C + A} \\
\midrule
Appeal to Emotion & \textbf{44\%} & 38\% & 41\% \\
Appeal to Authority & \textbf{49\%} & 36\% & 47\% \\
Ad Hominem & 73\% & \textbf{77\%} & 72\% \\
False Cause & 45\% & \textbf{51\%} & 31\% \\
Slippery Slope & \textbf{50\%} & 38\% & 32\% \\
Slogans & \textbf{63\%} & 52\% & 52\% \\
\bottomrule
\end{tabular}
\caption{Per-class F1-score across input conditions. Bold values highlight the highest score per fallacy type.}
\label{tab:pd-f1-per-class}
\end{table}

\begin{table}[H]
\centering
\begin{tabular}{lccc}
\toprule
\textbf{Fallacy Type} & \textbf{P-B} & \textbf{P-C} & \textbf{P-C + A} \\
\midrule
Appeal to Emotion & \textbf{33\%} & 28\% & 27\% \\
Appeal to Authority & 53\% & 46\% & \textbf{70\%} \\
Ad Hominem & 71\% & 71\% & \textbf{74\%} \\
False Cause & 45\% & \textbf{48\%} & 42\% \\
Slippery Slope & \textbf{88\%} & 83\% & 80\% \\
Slogans & 73\% & 73\% & \textbf{100\%} \\
\bottomrule
\end{tabular}
\caption{Per-class Precision across input conditions. Bold values highlight the highest score per fallacy type.}
\label{tab:pd-precision-per-class}
\end{table}

\begin{table}[H]
\centering
\begin{tabular}{lccc}
\toprule
\textbf{Fallacy Type} & \textbf{R-B} & \textbf{R-C} & \textbf{R-C + A} \\
\midrule
Appeal to Emotion & 65\% & 60\% & \textbf{90\%} \\
Appeal to Authority & \textbf{45\%} & 30\% & 35\% \\
Ad Hominem & 75\% & \textbf{85\%} & 70\% \\
False Cause & 45\% & \textbf{55\%} & 25\% \\
Slippery Slope & \textbf{35\%} & 25\% & 20\% \\
Slogans & \textbf{55\%} & 40\% & 35\% \\
\bottomrule
\end{tabular}
\caption{Per-class Recall across input conditions. Bold values highlight the highest score per fallacy type.}
\label{tab:pd-recall-per-class}
\end{table}

Tables~\ref{tab:pd-f1-per-class}, \ref{tab:pd-precision-per-class}, and \ref{tab:pd-recall-per-class} present the per-class F1-scores, precision, and recall respectively for each fallacy type across the three input conditions. Overall, the base \textbf{(B)} condition  provides the most balanced performance across categories, especially for \textit{Slippery Slope}, where the biggest performance drop is seen in both the context \textbf{(C)} and context + audio \textbf{(C + A)} conditions. The addition of context improves performance for reasoning related fallacies (where context is useful) such as \textit{Ad Hominem} and \textit{False Cause}, both of which show higher F1-Scores (77\% and 51\%) and improved recall (85\% and 55\%). However, the context + audio condition yields even worse results, while precision increases significantly for \textit{Appeal to Authority} (70\%) and \textit{Slogans} (100\%), this comes at the cost of recall in all categories except for \textit{Appeal to Emotion}. Notably, \textit{Appeal to Emotion} shows a very large increase in recall when audio input is included, aligning with its additional emotional tone metadata. However, as a result, F1-scores remain lower than both the base and context conditions due to reduced overall precision. These results suggest that context might potentially assist for logical fallacies, but emotional tone metadata may cause the model to overfit on that data, increasing the likelihood of misclassifying statements as \textit{Appeal to Emotion} and reducing overall performance.

\subsubsection{Confusion Matrices}
\begin{figure}[H]
    \centering
    \includegraphics[width=0.9\linewidth]{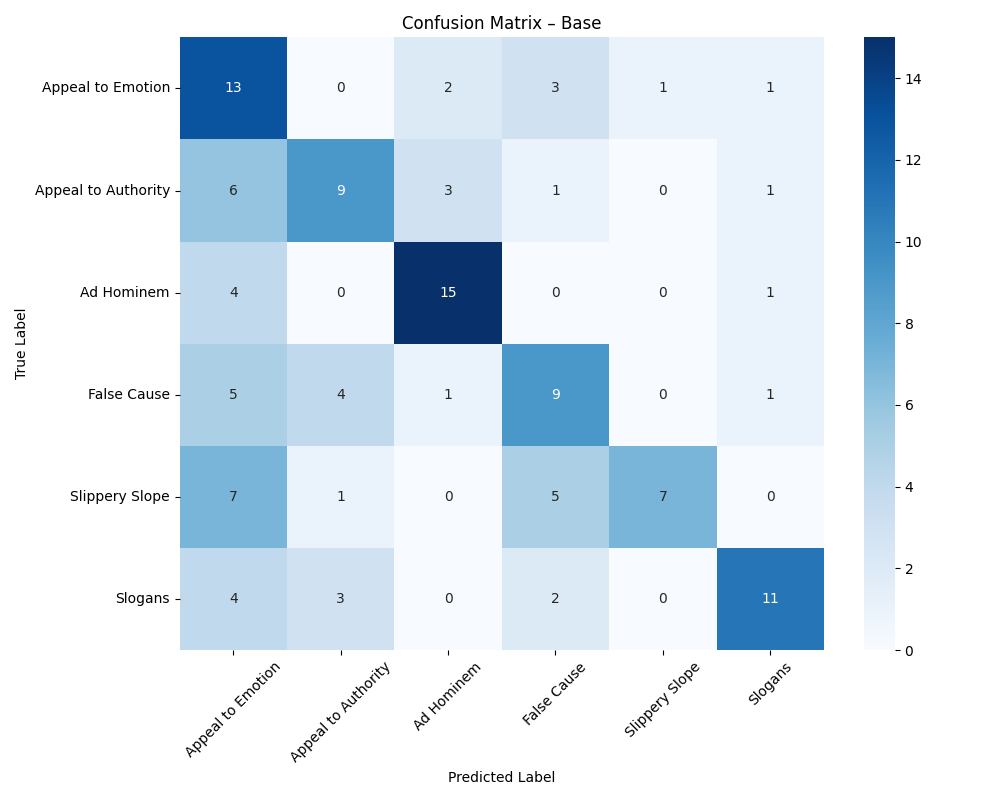}
    \caption{Confusion matrix for the Base condition under the PD prompt.}
    \label{fig:pd_cm_base}
\end{figure}

\begin{figure}[H]
    \centering
    \includegraphics[width=0.9\linewidth]{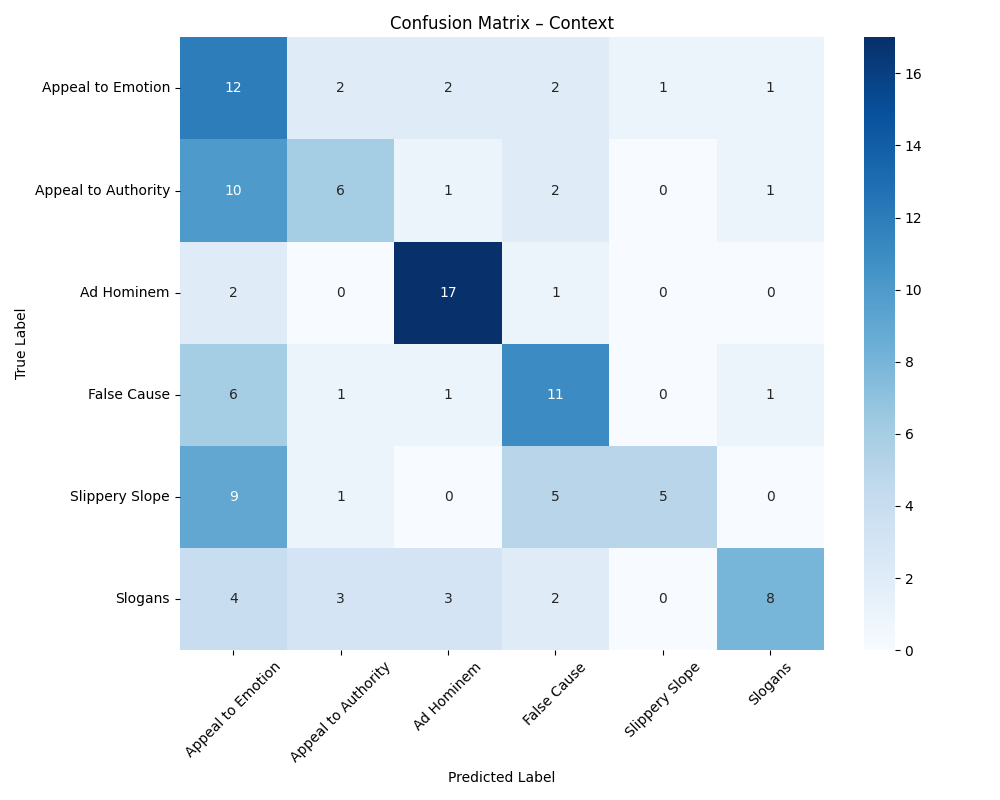}
    \caption{Confusion matrix for the Context condition under the PD prompt.}
    \label{fig:pd_cm_context}
\end{figure}

\begin{figure}[H]
    \centering
    \includegraphics[width=0.9\linewidth]{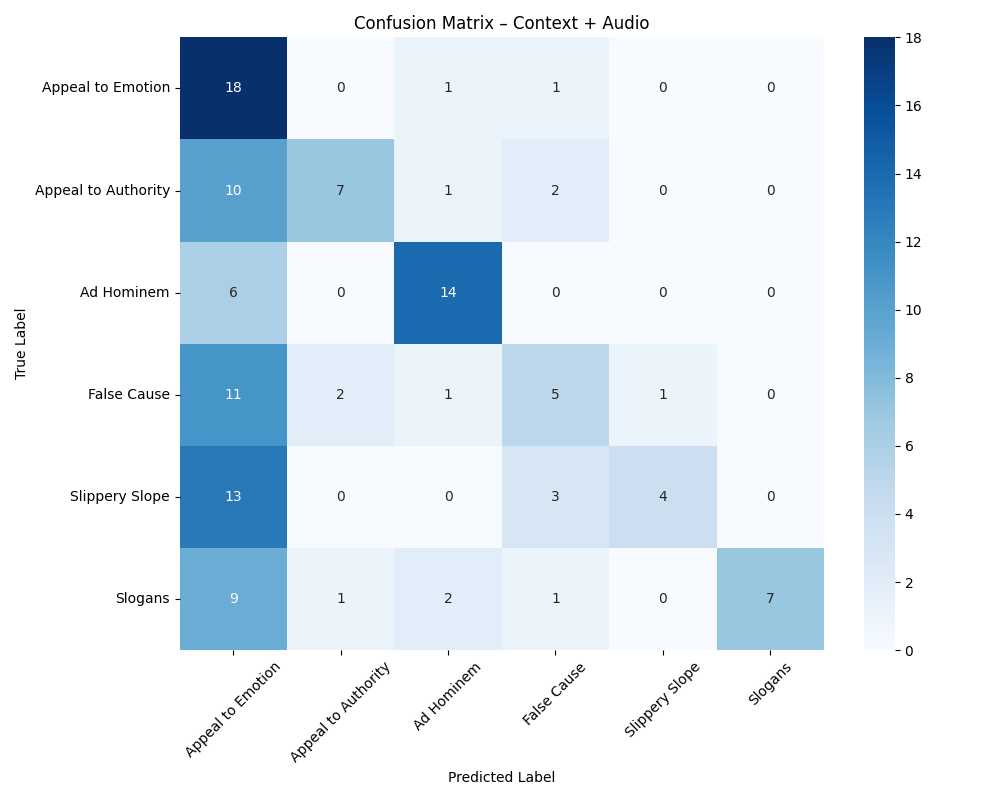}
    \caption{Confusion matrix for the Context + Audio condition under the PD prompt.}
    \label{fig:pd_cm_audio}
\end{figure}

The base confusion matrix in Figure~\ref{fig:pd_cm_base} reflects a relatively balanced classification baseline, though it slightly underperforms on \textit{Slippery Slope} and often misclassifies statements as \textit{Appeal to Emotion}. Figure~\ref{fig:pd_cm_context}, which shows the context-only condition, reveals improved classification for \textit{Ad Hominem} and \textit{False Cause}, but introduces confusion between \textit{Appeal to Emotion} and \textit{Appeal to Authority}, suggesting that contextual information may blur the line between an \textit{Appeal to Emotion} or \textit{Authority}. Figure~\ref{fig:pd_cm_audio}, representing the context + audio condition, shows an even larger shift toward \textit{Appeal to Emotion}, at the expense of overall performance.

\begin{figure}[H]
    \centering
    \includegraphics[width=0.9\linewidth]{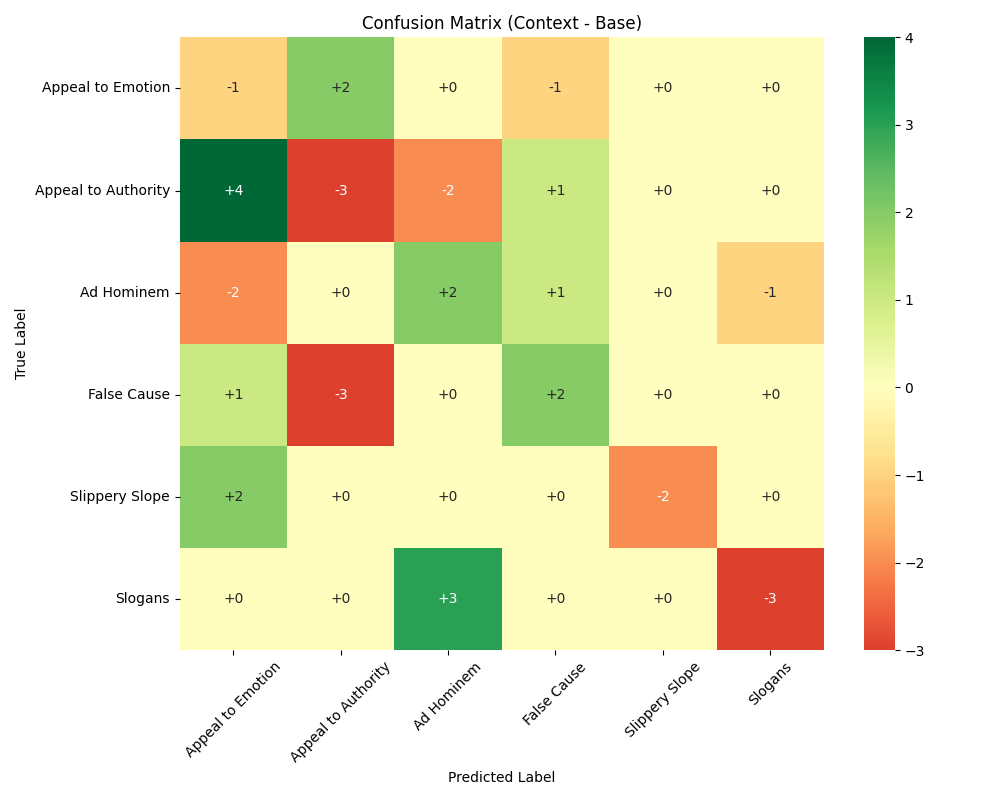}
    \caption{Confusion matrix for the \textit{Context condition \(-\) Base condition} under the PD prompt.}
    \label{fig:pd_cm_diff_context}
\end{figure}

\begin{figure}[H]
    \centering
    \includegraphics[width=0.9\linewidth]{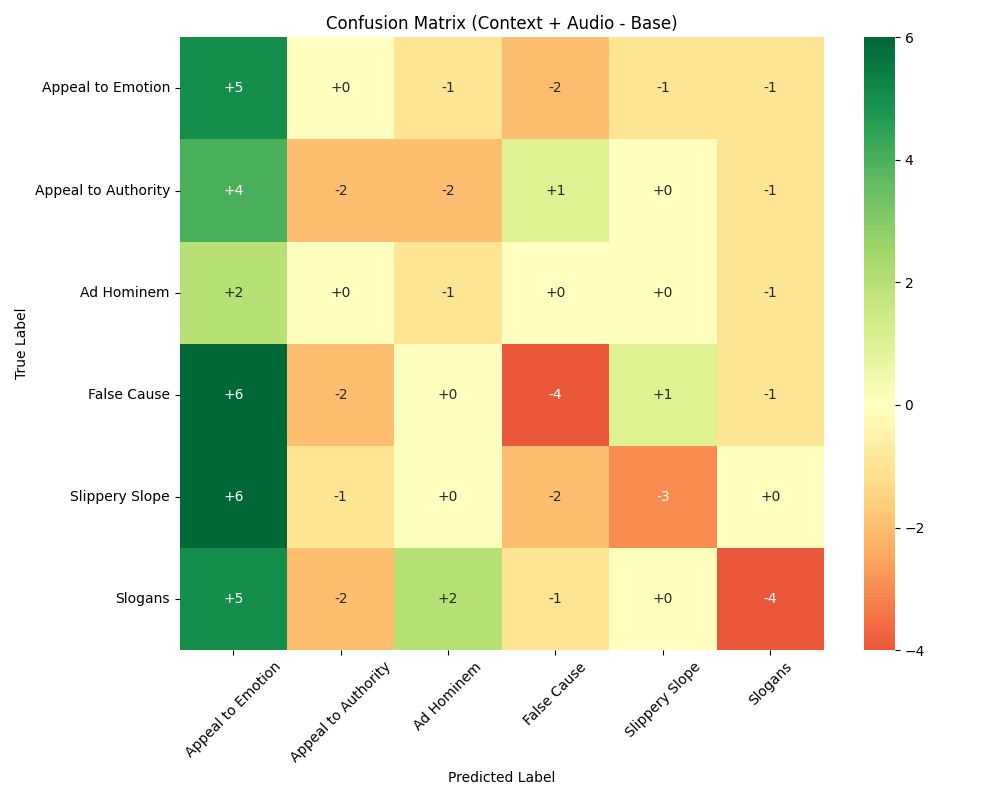}
    \caption{Difference matrix for the \textit{Context + Audio condition \(-\) Base condition} under the PD prompt.}
    \label{fig:pd_cm_diff_audio}
\end{figure}

To further illustrate these shifts, Figures~\ref{fig:pd_cm_diff_context} and~\ref{fig:pd_cm_diff_audio} present difference matrices that quantify changes in model behavior when compared to the base condition. In Figure~\ref{fig:pd_cm_diff_context}, the \textit{Context \(-\) Base} matrix shows small improvements (+2) in correctly classified statements of \textit{False Cause} and \textit{Appeal to Authority}, as well as, increased misclassifications between \textit{Appeal to Emotion} and \textit{Appeal to Authority}. Figure~\ref{fig:pd_cm_diff_audio} representing the \textit{Context + Audio \(-\) Base} condition shows a preference for predicting \textit{Appeal to Emotion}, causing a large increase in misclassifications, indicating that emotional tone metadata biases the model toward this specific fallacy type.

\subsubsection{Qualitative Analysis}
To illustrate how the model's reasoning varied across input conditions, we analyzed two representative examples: 
\begin{enumerate}
    \item \textit{"She doesn’t have the look."} 
        
        The true fallacy label for this case is \textbf{Ad Hominem (2)} . However, the model predicted three different labels depending on the input condition:

        \begin{itemize}
            \item \textbf{Base}: \textit{Slogans} (5), “The closest category is Slogans (5) because it's a vague, loaded phrase.”
            \item \textbf{Context}: \textit{Ad Hominem} (2), “The attack on 'the look' is a form of name-calling, which falls under the Ad Hominem category.”
            \item \textbf{Context + Audio}: \textit{Appeal to Emotion} (0), “The statement uses emotionally charged language to sway the audience, which fits Appeal to Emotion.”
        \end{itemize}

        This example highlights how the base prompt focused primarily on surface level vagueness, the context allowed for correct classification of a personal attack, and the emotional tone metadata caused the model to focus too much on the emotional language, resulting in a misclassification.
        
    \item \textit{"And the 'Los Angeles Times' spent a lot of ink going into that."}

        The true fallacy label is \textbf{Appeal to Authority (1)}. The model produced three different classifications across the input conditions:

        \begin{itemize}
            \item \textbf{Base}: \textit{Ad Hominem} (2), “...undermining the opponent's position without addressing the actual standpoint. This aligns with Ad Hominem (2), as it attacks the source’s credibility rather than engaging with the argument’s merits.”
            \item \textbf{Context}: \textit{Appeal to Emotion} (0), “...which could be interpreted as an appeal to emotion (0) by leveraging public perception of media credibility…”
            \item \textbf{Context + Audio}: \textit{Appeal to Authority} (1), “This aligns with the Appeal to Authority (1) category, as the media outlet is implicitly treated as an authority…”
        \end{itemize}

        This example shows how each input condition changed the interpretation. The base model saw a personal attack (Ad Hominem), while context misread it as an emotional ploy (Appeal to Emotion). Only context + audio correctly classified it as an Appeal to Authority, recognizing the use of media credibility as support for the argument.
\end{enumerate}

\subsection{Periodic Table of Arguments}
The Periodic Table of Arguments (PTA) \cite{wagemansHowIdentifyArgument2023} prompt uses a structural constraint-based approach to fallacy detection. It is designed to analyse arguments through three parameters: Form, Substance, and Lever. Unlike the rule-based PD approach, PTA identifies fallacies as violations of valid argument patterns rather than conversational rules. This section evaluates how the PTA prompt performed across the same three input conditions and analyzes the model's adherence to structural reasoning principles.

\subsubsection{Performance Matrices}
Table \ref{tab:pta-metrics} presents the overall classification metrics for the PTA prompt across input conditions. The results reveal a notably different pattern compared to the pragma-dialectical approach. The base condition achieved 44\% accuracy, while context slightly improved performance to 45\%. However, the context + audio condition showed dramatic performance degradation, dropping to just 25\% accuracy across all metrics.

The macro precision follows an inverse pattern. It is shown that the base condition achieving the highest score at 69\%, declining to 58\% with context, and collapsing to 25\% with audio input.This suggests that while the base PTA prompt maintains relatively precise classifications, additional contextual information possibly disrupts the model's structural reasoning process more severely than observed with the PD approach. The uniform degradation across all metrics in the context + audio condition indicates a fundamental breakdown in the model's ability to apply PTA constraints when emotional metadata is present.

\begin{table}[H]
\centering
\begin{tabular}{lccc}
\toprule
\textbf{Metric} & \textbf{B} & \textbf{C} & \textbf{C + A} \\
\midrule
Accuracy & 44\% & \textbf{45\%} & 25\% \\
Macro Precision & \textbf{69\%} & 58\% & 25\% \\
Macro Recall & 44\% & \textbf{45\%} & 25\% \\
Macro F1 & 43\% & \textbf{44\%} & 25\% \\
Weighted F1 & 43\% & \textbf{44\%} & 22\% \\
\bottomrule
\end{tabular}
\caption{Overall classification metrics for the PTA prompt across input conditions. Bold values indicate the highest score per row.}
\label{tab:pta-metrics}
\end{table}

\subsubsection{Per-Class Performance}
Tables \ref{tab:pta-f1-per-class}, \ref{tab:pta-precision-per-class}, and \ref{tab:pta-recall-per-class} present the per-class performance metrics across input condistions. The F1-scores reveal that Slogans consistently performs best across all conditions (55\%, 52\%, 30\%), followed by Appeal to Emotion and Appeal to Authority in base and context conditions. Ad Hominem shows an interesting pattern, with poor performance in the base condition (33\%) but significant improvement with context (62\%), suggesting that PTA's structural analysis benefits from additional contextual information for personal attack identification.

The precision scores highlight extreme patterns for Ad Hominem and Slippery Slope. They achieve 100\% precision in the base condition but with correspondingly low recall. This indicates that when the model correctly identifies these fallacies, but misses many instances. Slippery Slope shows the most dramatic degradation, maintaining reasonable precision in base and context conditions but completely failing in the context + audio condition (0\% across all metrics). 

According to the recall patterns, False Cause is the most stable performance across all conditions (75\%, 75\%, 65\%). Meanwhile, other types of fallacy experience substantial drops with audio input. Appeal to Authority and Slogans show severe recall degradation in the context + audio condition (5\% and 2\% respectively), suggesting that emotional metadata fundamentally disrupts the model's ability to recognize these structural patterns. This suggests that PTA's constraint-based reasoning is more sensitive to emotional interference than the rule-based PD approach.

\begin{table}[H]
\centering
\begin{tabular}{lccc}
\toprule
\textbf{Fallacy Type} & \textbf{F1-B} & \textbf{F1-C} & \textbf{F1-C + A} \\
\midrule
Appeal to Emotion & \textbf{51\%} & 48\% & 18\% \\
Appeal to Authority & \textbf{51\%} & 47\% & 7\% \\
Ad Hominem & 33\% & \textbf{62\%} & 39\% \\
False Cause & 41\% & \textbf{43\%} & 37\% \\
Slippery Slope & \textbf{26\%} & 9\% & 0\% \\
Slogans & \textbf{55\%} & 52\% & 30\% \\
\bottomrule
\end{tabular}
\caption{Per-class F1-score across input conditions. Bold values highlight the highest score per fallacy type.}
\label{tab:pta-f1-per-class}
\end{table}

\begin{table}[H]
\centering
\begin{tabular}{lccc}
\toprule
\textbf{Fallacy Type} & \textbf{P-B} & \textbf{P-C} & \textbf{P-C + A} \\
\midrule
Appeal to Emotion & \textbf{40\%} & 38\% & 14\% \\
Appeal to Authority & \textbf{60\% }& 57\% & 11\% \\
Ad Hominem & \textbf{100\%} & 83\% & 44\% \\
False Cause & 28\% & \textbf{31\%} & 25\% \\
Slippery Slope & \textbf{100\%} & 50\% & 0\% \\
Slogans & \textbf{89\%} & 88\% & 57\% \\
\bottomrule
\end{tabular}
\caption{Per-class Precision across input conditions. Bold values highlight the highest score per fallacy type.}
\label{tab:pta-precision-per-class}
\end{table}

\begin{table}[H]
\centering
\begin{tabular}{lccc}
\toprule
\textbf{Fallacy Type} & \textbf{R-B} & \textbf{R-C} & \textbf{R-C + A} \\
\midrule
Appeal to Emotion & \textbf{70\%} & 65\% & 25\% \\
Appeal to Authority & \textbf{45\%} & 40\% & 5\% \\
Ad Hominem & 20\% & \textbf{50\%} & 35\% \\
False Cause & \textbf{75\%} & \textbf{75\%} & 65\% \\
Slippery Slope & \textbf{15\%} & 5\% & 0\% \\
Slogans & \textbf{40\%} & 37\% & 2\% \\
\bottomrule
\end{tabular}
\caption{Per-class Recall across input conditions. Bold values highlight the highest score per fallacy type.}
\label{tab:pta-recall-per-class}
\end{table}

\subsubsection{Confusion Matrices}
\begin{figure}[H]
    \centering
    \includegraphics[width=0.9\linewidth]{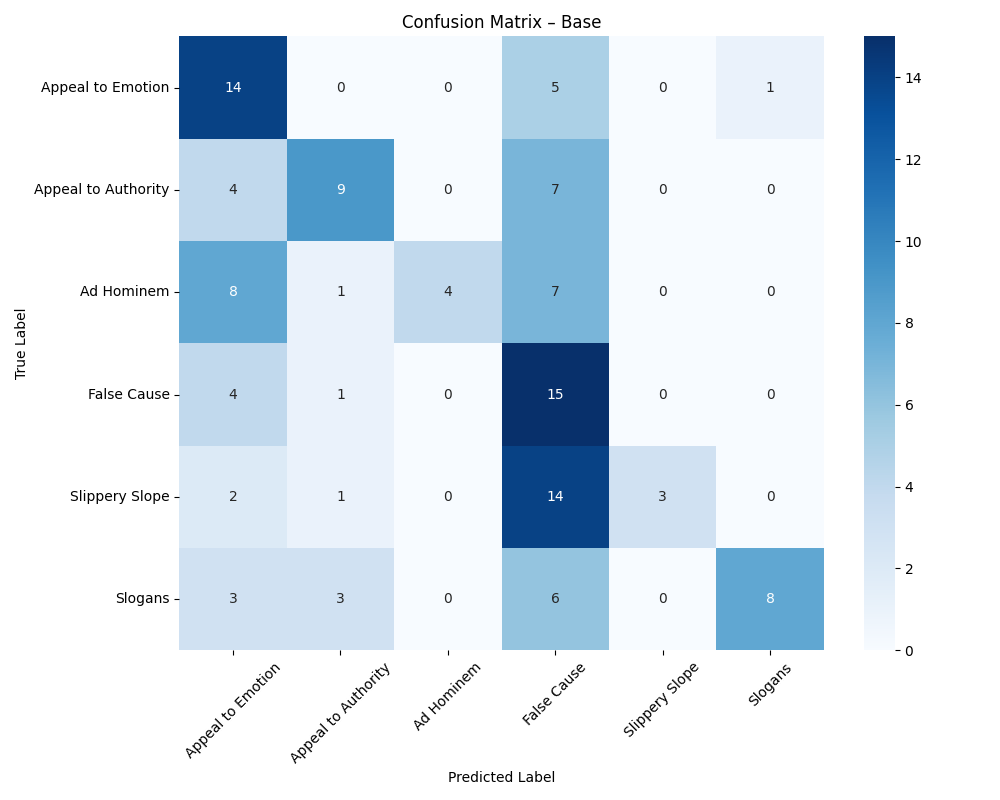}
    \caption{Confusion matrix for the Base condition under the PTA prompt.}
    \label{fig:pta_cm_base}
\end{figure}

\begin{figure}[H]
    \centering
    \includegraphics[width=0.9\linewidth]{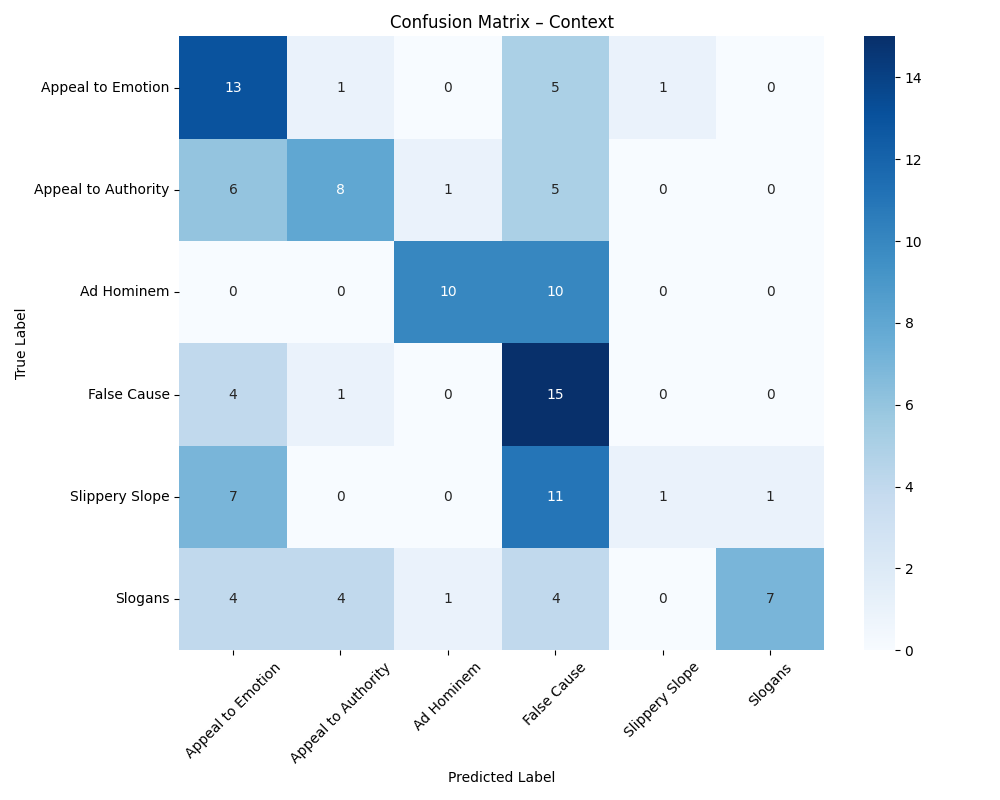}
    \caption{Confusion matrix for the Context condition under the PTA prompt.}
    \label{fig:pta_cm_context}
\end{figure}

\begin{figure}[H]
    \centering
    \includegraphics[width=0.9\linewidth]{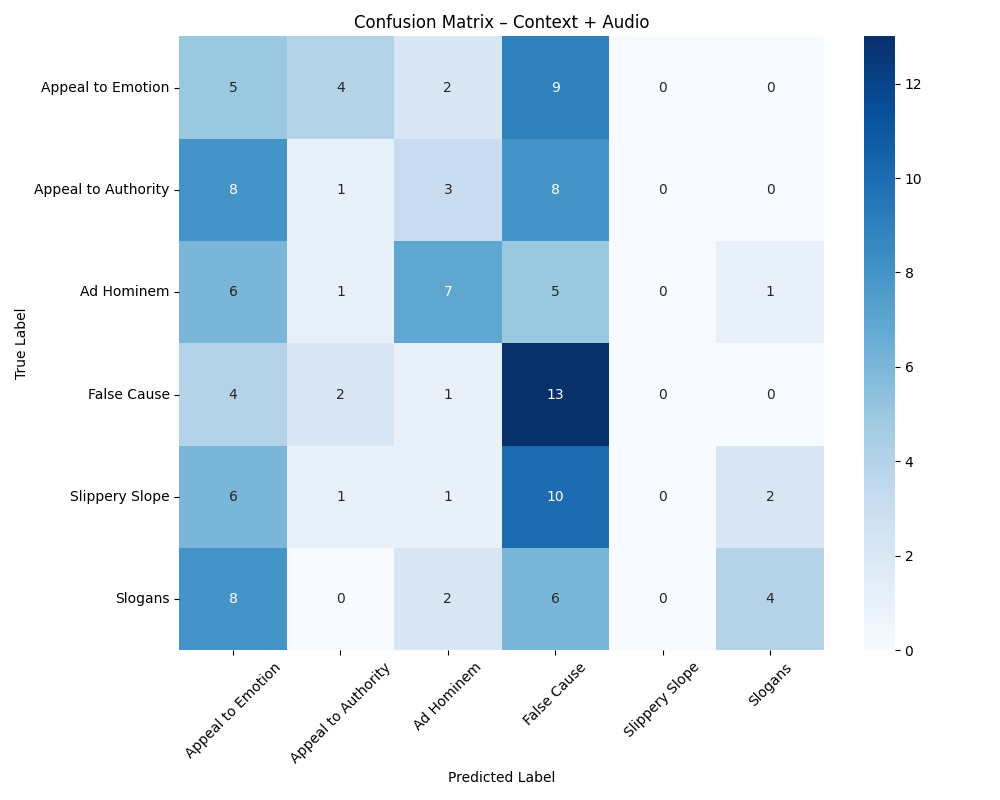}
    \caption{Confusion matrix for the Context + Audio condition under the PTA prompt.}
    \label{fig:pta_cm_all}
\end{figure}

The PTA confusion matrices in Figure \ref{fig:pta_cm_base}, \ref{fig:pta_cm_context}
, and \ref{fig:pta_cm_all} reveal distinct patterns compared to the pragma-dialectical approach. The base condition (Figure \ref{fig:pta_cm_base}) maintains balanced performance despite notable Slippery Slope weakness and Slogans overprediction, with more concentrated predictions than PD's distributed pattern. Adding context (Figure \ref{fig:pta_cm_context}) improves Ad Hominem classification but creates confusion between Appeal to Authority and other categories. While PD shows general contextual benefits, PTA specifically enhances structural personal attack identification at the cost of authority-based reasoning clarity. The context + audio condition produces dramatic breakdown (Figure \ref{fig:pta_cm_all}), concentrating predictions toward Appeal to Emotion with severe misclassification across other categories. This degradation exceeds that observed in PD matrices, suggesting heightened vulnerability of constraint-based reasoning to emotional interference.

\subsubsection{Qualitative Analysis}
To maintain controlled comparison with the pragma-dialectical analysis, we examine the same two examples to understand how PTA's structural constraint framework performs relative to PD's rule-based approach.

\begin{enumerate}
    \item \textit{She doesn't have the look.}
    
    The true fallacy label is Ad Hominem (2). PTA's classifications show a different trajectory than PD. The base condition incorrectly identified Appeal to Emotion (0). Its reasoning focused on missing levers rather than personal attack patterns. However, context enabled correct classification as Ad Hominem (2), with the model successfully applying structural analysis: "attacking the opponent's stamina, which is a personal attribute, so it's an ad hominem." Notably, PTA maintained this correct classification even with audio input, unlike PD which shifted to Appeal to Emotion under emotional influence.

    \item \textit{And the "Los Angeles Times" spent a lot of ink going into that.}
    
    The true fallacy label is Appeal to Authority (1). Here, PTA demonstrated superior base performance compared to PD, correctly identifying the structural violation with detailed form-substance-lever analysis. The reasoning traced through Alpha form identification, Fact substance recognition, and lever evaluation, concluding with proper Appeal to Authority classification. Context maintained this accuracy, but audio input caused degradation to Appeal to Emotion (0), reflecting the general pattern of emotional metadata disrupting structural reasoning.
\end{enumerate}

These examples demonstrate core differences between the frameworks. PTA's constraint-based approach shows particular strength in authority-related fallacies where structural analysis can identify inappropriate levers, but struggles with personal attacks when contextual grounding is absent. The reasoning traces reveal that PTA prompts encourage systematic theoretical application, though this structured approach appears more vulnerable to emotional interference than PD's rule-based method. However, given the limited scope of these illustrative cases, broader patterns would require examination of the full dataset to validate these preliminary observations.

\section{Conclusions and Discussion}
This study explored how the addition of context and emotional tone metadata influenced large language model (LLM) performance in fallacy classification. Furthermore, we experimented with theoretically grounded chain-of-thought (CoT) prompting, drawing from Pragma-Dialectics and the Periodic Table of Arguments, to analyze whether structured reasoning frameworks could potentially enhance model interpretability and accuracy. Our results show that the inclusion of textual context improved the model's ability to reason for specific fallacy types that depend on context. However, this benefit came at the cost of decreased overall performance.

Interestingly, the inclusion of emotional tone metadata (audio-based context) introduced a clear shift in classification behavior. Specifically, models became more likely to label arguments as \textit{Appeal to Emotion}, even when the actual fallacy was structurally or logically driven. This suggests that emotional tone metadata can dominate the model’s reasoning, causing it to overfit on emotional features at the expense of logical reasoning.

The base prompts (lacking any added context or emotion) outperformed the enhanced versions. This suggests that relatively basic inputs may help the model focus on the core of the statement without distraction. Rather than improving performance, additional context mostly diluted or diverted the model’s attention.

A comparison of results across prompt types shows that base prompts often performed as well as or better than versions with added context or emotional tone metadata. Context helped in some cases, but these gains were inconsistent. Emotional tone metadata often reduced performance, especially in the PTA prompts. Of all the combinations, the highest overall performance was achieved using the base prompt with the Pragma-Dialectics framework. This indicates that the theoretical grounding provided a slight, but not substantial, boost in performance, specifically for the Pragma-Dialectics framework. \textit{Ad Hominem} consistently achieved the highest scores, while \textit{Slippery Slope} remained the most difficult to classify. These results reinforce that the usefulness of additional context depends on both the fallacy type and the prompting framework.

This also highlights the model’s sensitivity to prompt framing. The same statement, when packaged with different contextual information, can yield different predictions depending on which aspect of the prompt the model prioritizes. Overall, while additional context and emotional tone metadata offer potential benefits, they also introduce noise. When both context and audio are added, performance tends to degrade, likely due to the increased noise. This suggests that both the quality and integration method of additional context play a important role in model performance.

In summary, context and emotional tone metadata added in prompting can both improve and worsen LLM reasoning. A more tailored approach (designing prompts specific to each statement) could improve performance, but this would require a large manual effort and runs counter to the goal of leveraging LLMs for autonomous reasoning.

\subsection{Discussion}
Our findings suggest that additional context, intended to improve fallacy classification, instead worsened LLM performance. This suggests an idea of "attention dilution," where additional context or emotional tone metadata distracted the model from an core of the statement, leading to worse results than more basic prompts.

This effect was most noticed with emotional tone metadata, which biased the model toward labeling statements as \textit{Appeal to Emotion} at the cost of logical reasoning. Adding to this issue, the task's design introduced ambiguity by mapping PTA arguments and PD rules into six specific fallacy types, which likely added to the incorrect reasoning and resulting misclassifications.

This study's conclusions are preliminary, constrained by a few limitations:
\begin{itemize}
    \item Task Complexity: The model sometimes failed to follow the Chain-of-Thought (CoT) instructions, and some inputs from the dataset potentially contained multiple fallacies, making it even more difficult for the model to classify a single fallacy label.
    \item Experimental Scope: The use of a single model (Qwen3-8B), a small dataset, and a lack of temperature setting experimentation, limits the generalizability of our results.
\end{itemize}

These results highlight that a \textit{one-size-fits-all} approach to context is suboptimal for fallacy classification using LLMs.

\bibliographystyle{chicago}
\bibliography{mybib.bib}

\begin{thebibliography}{}

\bibitem[\protect\citeauthoryear{Goffredo, Haddadan, Vorakitphan, Cabrio, and Villata}{Goffredo et~al.}{2022}]{ijcai2022p575}
Goffredo, P., S.~Haddadan, V.~Vorakitphan, E.~Cabrio, and S.~Villata (2022, 7).
\newblock Fallacious argument classification in political debates.
\newblock In L.~D. Raedt (Ed.), {\em Proceedings of the Thirty-First International Joint Conference on Artificial Intelligence, {IJCAI-22}}, pp.\  4143--4149. International Joint Conferences on Artificial Intelligence Organization.
\newblock Main Track.

\bibitem[\protect\citeauthoryear{Habernal, Hannemann, Pollak, Klamm, Pauli, and Gurevych}{Habernal et~al.}{2017}]{habernal-etal-2017-argotario}
Habernal, I., R.~Hannemann, C.~Pollak, C.~Klamm, P.~Pauli, and I.~Gurevych (2017, September).
\newblock {A}rgotario: Computational argumentation meets serious games.
\newblock In L.~Specia, M.~Post, and M.~Paul (Eds.), {\em Proceedings of the 2017 Conference on Empirical Methods in Natural Language Processing: System Demonstrations}, Copenhagen, Denmark, pp.\  7--12. Association for Computational Linguistics.

\bibitem[\protect\citeauthoryear{Ifan}{Ifan}{2022}]{Ifan_2022}
Ifan, M. (2022, May).
\newblock Language as a weapon: The logical fallacies in political discourse.
\newblock {\em Journal of Society Innovation and Development\/}~{\em 3\/}(2), 18–27.

\bibitem[\protect\citeauthoryear{Liang, Tu, Du, and Xu}{Liang et~al.}{2024}]{liang2024multi}
Liang, X., G.~Tu, J.~Du, and R.~Xu (2024).
\newblock Multi-modal attentive prompt learning for few-shot emotion recognition in conversations.
\newblock {\em Journal of Artificial Intelligence Research\/}~{\em 79}, 825--863.

\bibitem[\protect\citeauthoryear{Lotfian and Busso}{Lotfian and Busso}{2017}]{lotfian2017building}
Lotfian, R. and C.~Busso (2017).
\newblock Building naturalistic emotionally balanced speech corpus by retrieving emotional speech from existing podcast recordings.
\newblock {\em IEEE Transactions on Affective Computing\/}~{\em 10\/}(4), 471--483.

\bibitem[\protect\citeauthoryear{Mancini, Ruggeri, and Torroni}{Mancini et~al.}{2024a}]{mancini-etal-2024-multimodal}
Mancini, E., F.~Ruggeri, and P.~Torroni (2024a, March).
\newblock Multimodal fallacy classification in political debates.
\newblock In Y.~Graham and M.~Purver (Eds.), {\em Proceedings of the 18th Conference of the European Chapter of the Association for Computational Linguistics (Volume 2: Short Papers)}, St. Julian{'}s, Malta, pp.\  170--178. Association for Computational Linguistics.

\bibitem[\protect\citeauthoryear{Mancini, Ruggeri, and Torroni}{Mancini et~al.}{2024b}]{manciniMultimodalFallacyClassification2024}
Mancini, E., F.~Ruggeri, and P.~Torroni (2024b).
\newblock Multimodal fallacy classification in political debates.
\newblock pp.\  170--178. Association for Computational Linguistics.

\bibitem[\protect\citeauthoryear{Mestre, Middleton, Ryan, Gheasi, Norman, and Zhu}{Mestre et~al.}{2023}]{mestre2023augmenting}
Mestre, R., S.~E. Middleton, M.~Ryan, M.~Gheasi, T.~Norman, and J.~Zhu (2023).
\newblock Augmenting pre-trained language models with audio feature embedding for argumentation mining in political debates.
\newblock In {\em Findings of the Association for Computational Linguistics: EACL 2023}, pp.\  274--288.

\bibitem[\protect\citeauthoryear{Mouchel, Paul, Cui, West, Bosselut, and Faltings}{Mouchel et~al.}{2025}]{mouchel2025logicalfallacyinformedframeworkargument}
Mouchel, L., D.~Paul, S.~Cui, R.~West, A.~Bosselut, and B.~Faltings (2025).
\newblock A logical fallacy-informed framework for argument generation.

\bibitem[\protect\citeauthoryear{Musi and Reed}{Musi and Reed}{2022}]{Elena}
Musi, E. and C.~Reed (2022).
\newblock From fallacies to semi-fake news: Improving the identification of misinformation triggers across digital media.
\newblock {\em Discourse \& Society\/}~{\em 33\/}(3), 349--370.

\bibitem[\protect\citeauthoryear{Tomar, Tiwari, Saha, and Saha}{Tomar et~al.}{2023}]{tomar2023your}
Tomar, M., A.~Tiwari, T.~Saha, and S.~Saha (2023).
\newblock Your tone speaks louder than your face! modality order infused multi-modal sarcasm detection.
\newblock In {\em Proceedings of the 31st ACM International Conference on Multimedia}, pp.\  3926--3933.

\bibitem[\protect\citeauthoryear{Van~Eemeren and Grootendorst}{Van~Eemeren and Grootendorst}{1987}]{vaneemerenFallaciesPragmadialecticalPerspective1987}
Van~Eemeren, F.~H. and R.~Grootendorst (1987, September).
\newblock Fallacies in pragma-dialectical perspective.
\newblock {\em Argumentation\/}~{\em 1\/}(3), 283--301.

\bibitem[\protect\citeauthoryear{Wagemans}{Wagemans}{2023}]{wagemansHowIdentifyArgument2023}
Wagemans, J.~H. (2023, January).
\newblock How to identify an argument type? {On} the hermeneutics of persuasive discourse.
\newblock {\em Journal of Pragmatics\/}~{\em 203}, 117--129.

\bibitem[\protect\citeauthoryear{Wagner, Triantafyllopoulos, Wierstorf, Schmitt, Burkhardt, Eyben, and Schuller}{Wagner et~al.}{2023}]{wagner2023dawn}
Wagner, J., A.~Triantafyllopoulos, H.~Wierstorf, M.~Schmitt, F.~Burkhardt, F.~Eyben, and B.~W. Schuller (2023).
\newblock Dawn of the transformer era in speech emotion recognition: closing the valence gap.
\newblock {\em IEEE Transactions on Pattern Analysis and Machine Intelligence\/}~{\em 45\/}(9), 10745--10759.

\bibitem[\protect\citeauthoryear{Wu, Zhang, and Li}{Wu et~al.}{2025}]{wu2025multi}
Wu, Y., S.~Zhang, and P.~Li (2025).
\newblock Multi-modal emotion recognition in conversation based on prompt learning with text-audio fusion features.
\newblock {\em Scientific Reports\/}~{\em 15\/}(1), 8855.

\bibitem[\protect\citeauthoryear{Zhao, Li, Jin, Wang, and Li}{Zhao et~al.}{2022}]{zhao2022memobert}
Zhao, J., R.~Li, Q.~Jin, X.~Wang, and H.~Li (2022).
\newblock Memobert: Pre-training model with prompt-based learning for multimodal emotion recognition.
\newblock In {\em ICASSP 2022-2022 IEEE International Conference on Acoustics, Speech and Signal Processing (ICASSP)}, pp.\  4703--4707. IEEE.

\end{thebibliography}

\section*{Appendix}


\subsection*{Dataset \& Statistics}

\subsection*{Audio Emotional Tone Feature Distribution}

\subsection*{Basic prompt \& Difference Matrices}
\begin{figure}[H]
    \centering
    \includegraphics[width=0.9\linewidth]{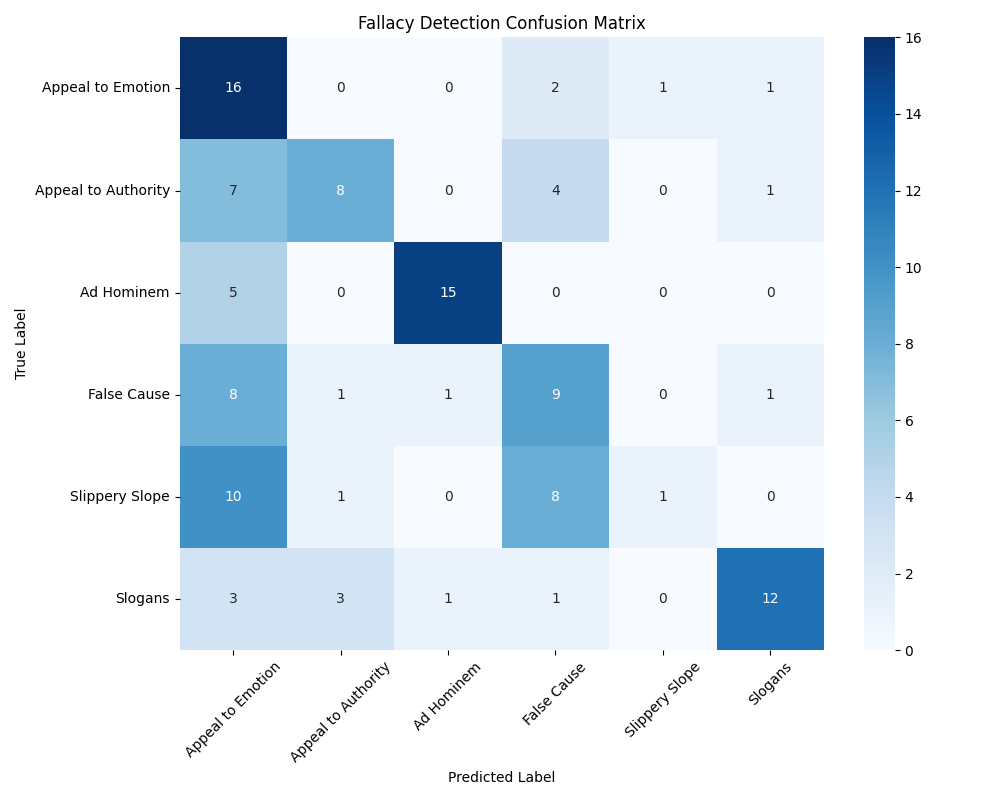}
    \caption{Confusion matrix for the Base condition under the basic prompt.}
    \label{fig:basic_cm_base}
\end{figure}

\begin{figure}[H]
    \centering
    \includegraphics[width=0.9\linewidth]{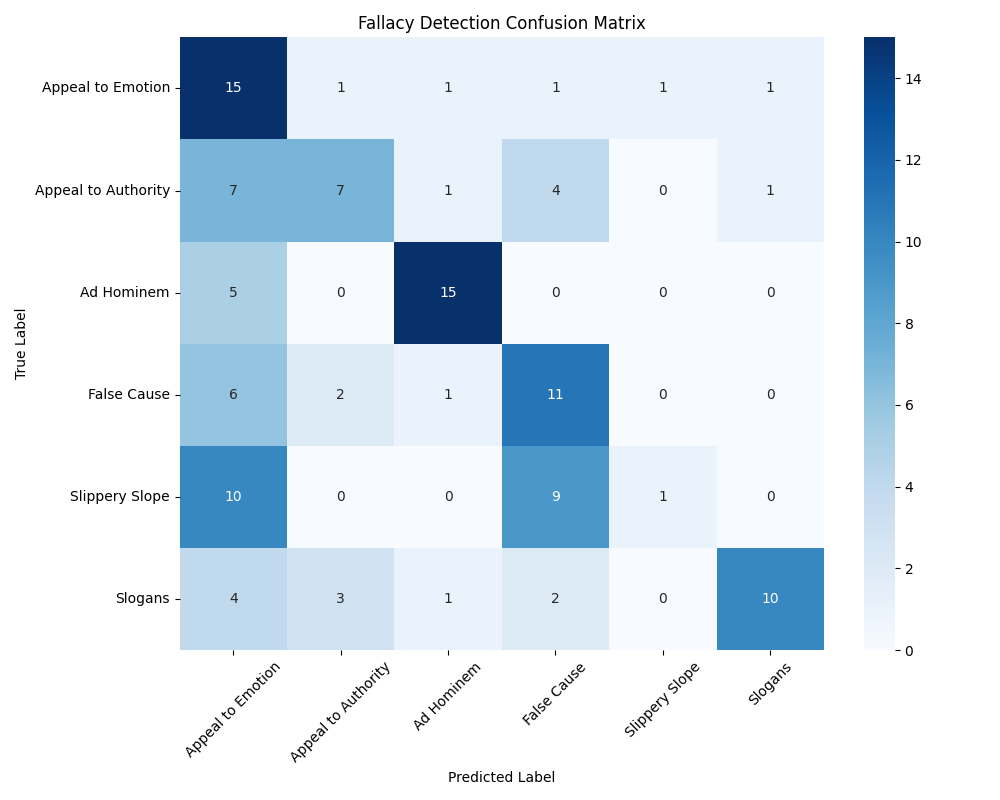}
    \caption{Confusion matrix for the Context condition under the basic prompt.}
    \label{fig:basic_cm_context}
\end{figure}

\begin{figure}[H]
    \centering
    \includegraphics[width=0.9\linewidth]{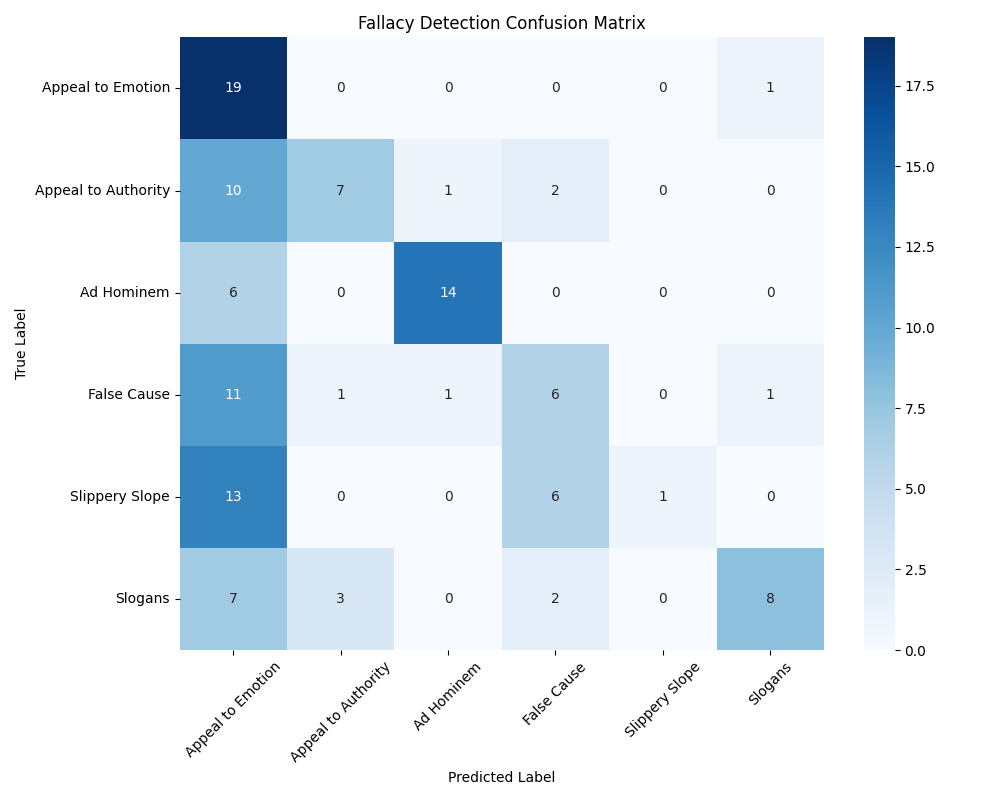}
    \caption{Confusion matrix for the Context + Audio condition under the basic prompt.}
    \label{fig:basic_cm_audio}
\end{figure}

\begin{figure}[H]
    \centering
    \includegraphics[width=0.9\linewidth]{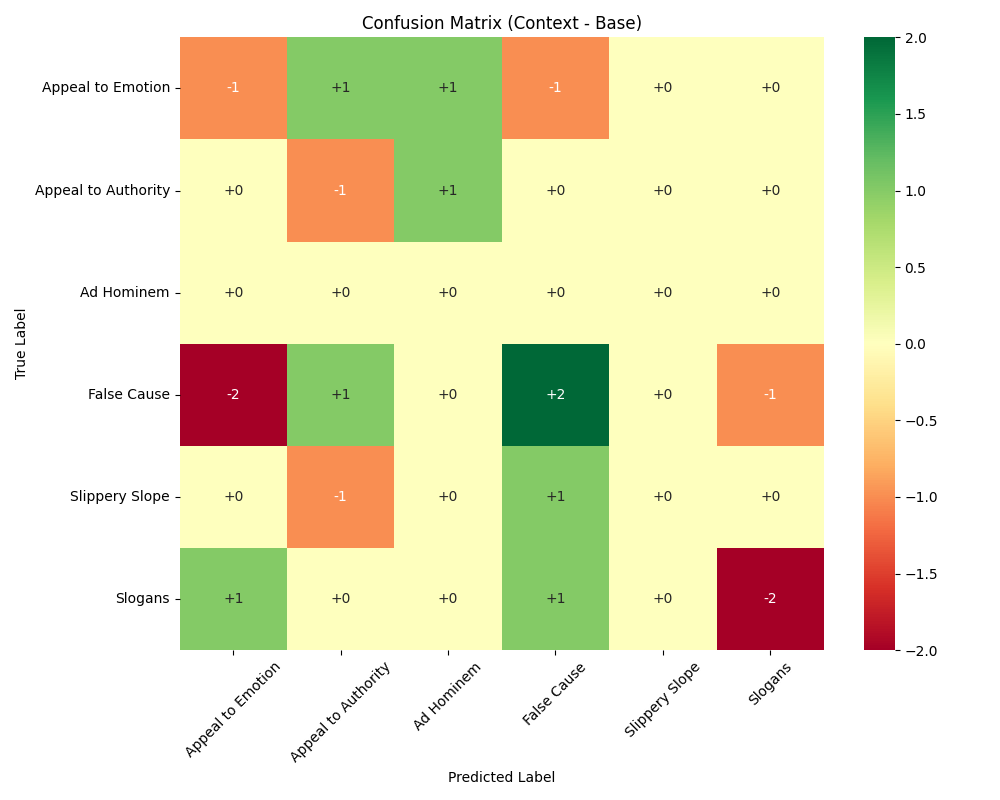}
    \caption{Difference matrix for the Context condition under the Basic prompt.}
    \label{fig:cm_diff_context}
\end{figure}

\begin{figure}[H]
    \centering
    \includegraphics[width=0.9\linewidth]{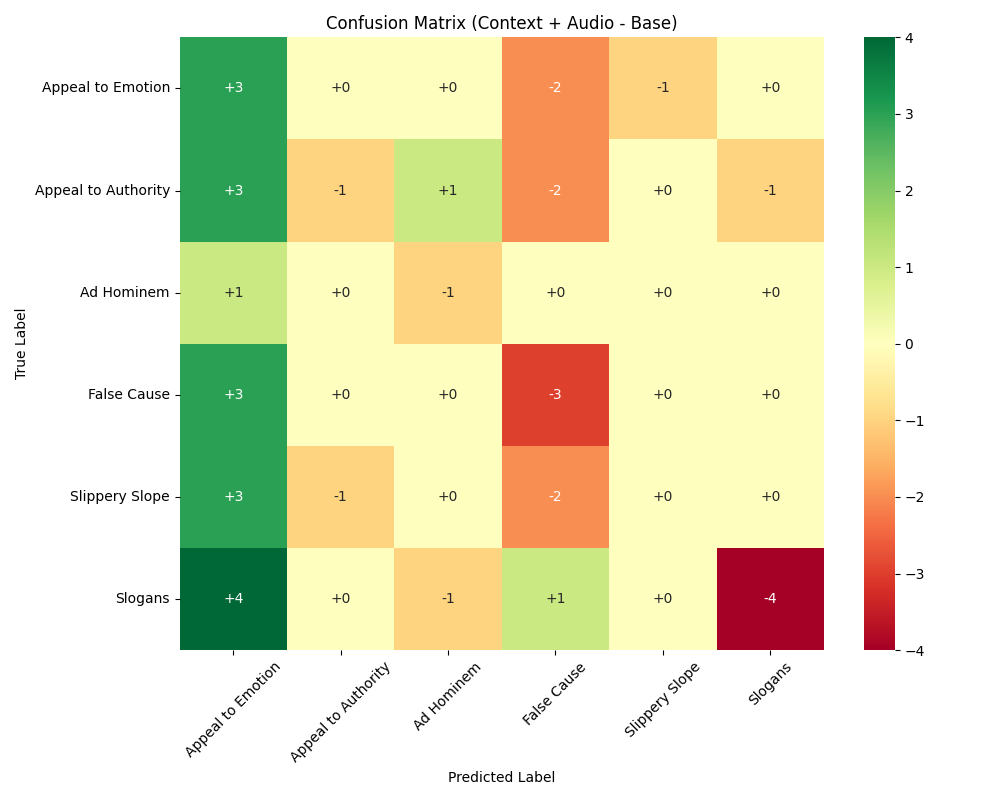}
    \caption{Difference matrix for the \textit{Context + Audio condition \(-\) Base condition} under the Basic prompt.}
    \label{fig:cm_diff_audio}
\end{figure}

\subsection*{Pragma-Dialectic Confusion \& Difference Matrices}

\subsection*{Periodic Table of Arguments Confusion \& Difference Matrices}

\begin{figure}[H]
    \centering
    \includegraphics[width=0.9\linewidth]{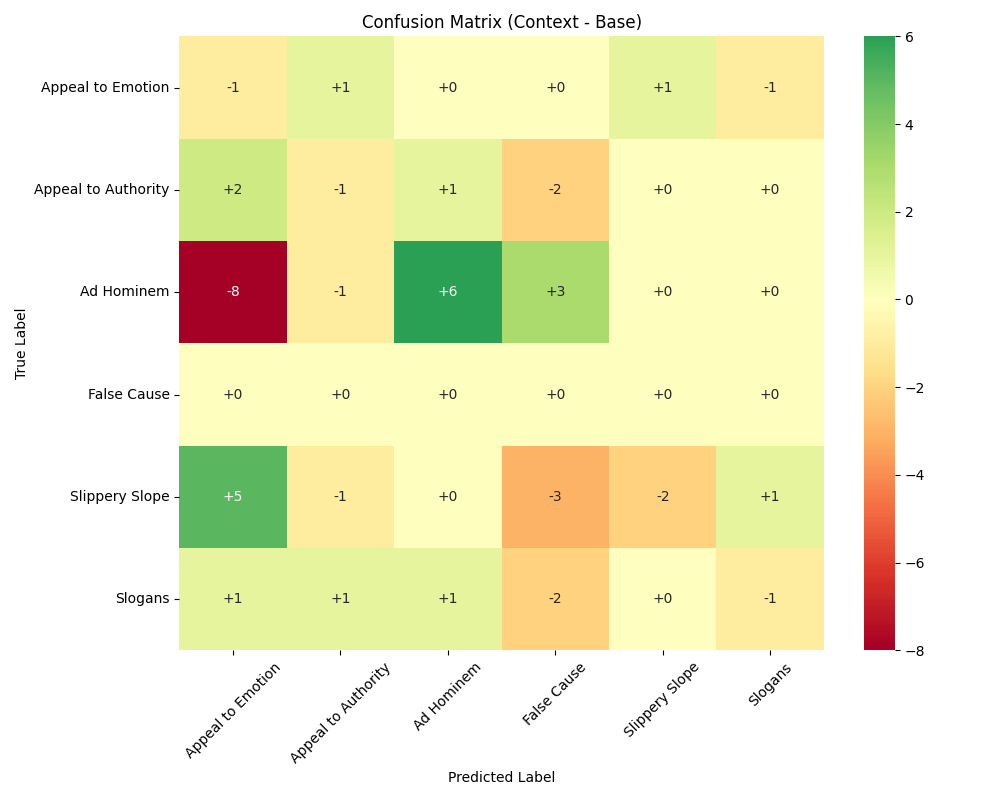}
    \caption{Difference matrix for the Context condition under the PTA prompt.}
    \label{fig:pta_dm_context}
\end{figure}

\begin{figure}[H]
    \centering
    \includegraphics[width=0.9\linewidth]{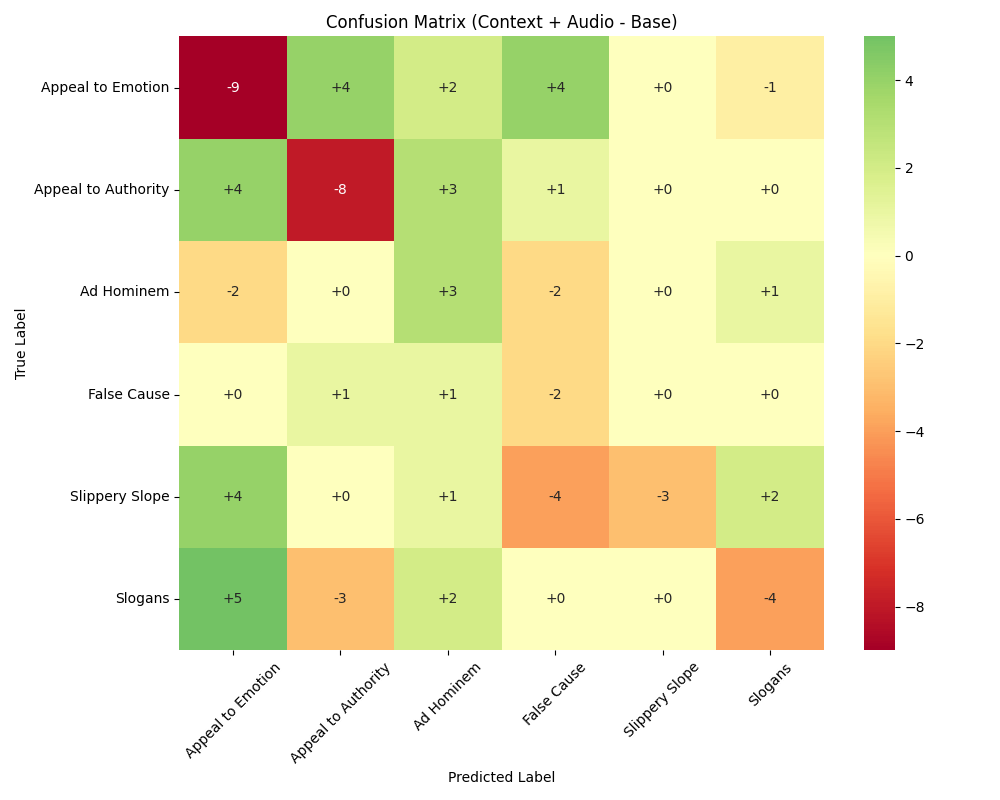}
    \caption{Difference matrix for the Context + Audio condition under the PTA prompt.}
    \label{fig:pta_dm_all}
\end{figure}
\end{document}